\documentclass{article}
\usepackage{spconf,amsmath,graphicx}
\usepackage{url}
\usepackage{subfigure}
\usepackage{amsmath}
\usepackage{amssymb}
\usepackage{amsfonts}
\newcommand{\argmin}{\mathop{\rm arg~min}\limits}
\usepackage{algorithm}
\usepackage{algpseudocode}
\usepackage{multirow}
\usepackage{appendix}
\algrenewcommand\algorithmicindent{1.0em}
\algnewcommand{\Inputs}[1]{%
  \State \textbf{Inputs:}
  \Statex \hspace*{\algorithmicindent}\parbox[t]{.8\linewidth}{\raggedright #1}
}
\algnewcommand{\Outputs}[1]{%
  \State \textbf{Outputs:}
  \Statex \hspace*{\algorithmicindent}\parbox[t]{.8\linewidth}{\raggedright #1}
}
\algnewcommand{\Initialize}[1]{%
  \State \textbf{Initialize:}
  \Statex \hspace*{\algorithmicindent}\parbox[t]{1.2\linewidth}{\raggedright #1}
}

\newcommand{\bp}{{\bf p}}

\renewcommand{\paragraph}[1]{\noindent\textbf{#1}\quad}

\usepackage{tabularx}
\usepackage{subfigure}

\makeatletter
\newcommand{\multiline}[1]{%
  \begin{tabularx}{\dimexpr\linewidth-\ALG@thistlm}[t]{@{}X@{}}
    #1
  \end{tabularx}
}
\makeatother

\title{Learning from Label Proportion with Online pseudo-label decision by Regret Minimization}
\name{Shinnosuke Matsuo, Ryoma Bise, Seiichi Uchida, Daiki Suehiro \vspace{-3mm}\thanks{This work was supported by JSPS KAKENHI Grant Number JP20H04211, JP21K12032, JP21K18312, JP22H05173, and JST ACT-X Grant Number JPMJAX200G, Japan.}}
\address{Kyushu University, Fukuoka, Japan}
\begin{document}
\ninept

\setlength{\abovedisplayskip}{3pt} 
\setlength{\belowdisplayskip}{3pt}
\maketitle

\begin{abstract}
This paper proposes a novel and efficient method for Learning from Label Proportions (LLP), whose goal is to train a classifier only by using the class label proportions of instance sets, called bags.
We propose a novel LLP method based on an online pseudo-labeling method with regret minimization. As opposed to the previous LLP methods, the proposed method effectively works even if the bag sizes are large.
We demonstrate the effectiveness of the proposed method using some benchmark datasets.
\end{abstract}

\section{Introduction}
\label{sec:intro}

Learning from Label Proportions (LLP)~\cite{quadrianto2009estimating,rueping2010svm} is a weakly-supervised machine learning task where only the class label proportion of the instances in each {\em bag} $B^i$ is given. A bag is a set of instances. Formally, for a $C$-class classification problem, multiple bags $B^1,\ldots, B^i, \ldots, B^n$ and the label proportion $\bp^i=(p^i_1,\ldots,p^i_c,\ldots,p^i_C)$ of each bag are given as the training set. For example, if $B^i$ contains 100, 50, and 50 instances of the class 1, 2, and 3, respectively, $\bp^i=(0.5, 0.25, 0.25)$. The goal of LLP is to train an instance classifier, just by the label proportion, that is, without the class label of each instance $x^i_j \in B^i$ ($j=1,\ldots,|B^i|$). Therefore, LLP is one of the most difficult weakly-supervised tasks.\par
%

Currently, the {\em proportion loss} is widely used for realizing LLP~\cite{ardehaly2017co,ijcai2021p377,liu2019learning}. It evaluates the difference between the given proportion $\bp^i$ and the proportion of the estimated labels of the $i$th bag $B^i$. 
However, it is known that the accuracy decreases for larger bags~\cite{liu2019learning,DulacArnoldG2020}. This weakness becomes crucial in many applications with large bags. An application example is a window-wise long-term signal classification with label proportion, where each signal is represented as a large bag with many instances corresponding to individual windows.

This paper proposes a new LLP method based on online pseudo-labeling by a regret minimization approach. 
In the proposed method, we assume a Deep Neural Network (DNN) as a classification model, and alternately update the model and pseudo labels along epochs. More precisely, at each $t$-th epoch, the DNN model is trained by the pseudo labels in a fully-supervised manner. Then the pseudo labels are updated by observing the behavior of the updated model. 
\par
One of the advantages of our method is that, by assigning the pseudo labels to the instances over the bags, we can make full use of instances to train a model even if the bag sizes are large. In other words, if we have $n$ instances, our method can train a model with $n$ instances with pseudo labels without depending on the bag sizes.
\par
Another advantage of our online pseudo-labeling approach is its strong theoretical support. Different from various heuristics-based pseudo-labeling approaches, ours follows the {\em regret} minimization framework, which is one of the theories for online decision-making. The regret is the difference between the actual decision and the best decision; in our case, the actual decision is the pseudo labels at each epoch, and the best decision is the best-performed pseudo labels averagely over the epochs. Our method has a theoretical upper bound of the regret --- this means that the performance of our method is not far away from the best-performed pseudo labels, although the pseudo labels are determined at each epoch in an online manner.  \par
To evaluate the performance of the proposed method, we use CIFAR10 for a synthetic LLP task. We observe how the proportion-loss-based methods perform with different sizes of bags and compare them with the proposed method. In addition, we conduct an ablation study to demonstrate the effectiveness of our pseudo-labeling approach based on regret minimization.

The main contributions of this paper are summarized as follows:
\begin{itemize}\setlength{\itemsep}{0pt}
    \item This paper proposes a novel and efficient LLP method, which can deal with even a very large bag.  
    \item The proposed method is based on online pseudo-labeling and has strong theoretical support in terms of regret minimization.
    \item The robustness to large bag sizes and the accuracy of the proposed method were validated through multiple comparative experiments using CIFAR-10 and SVHN. 
\end{itemize}

The code is publicly available at https://github.com/matsuo-shinnosuke/online-pseudo-labeling.

\section{Related Work}
\label{sec:related}
\noindent{\bf Learning from label proportions (LLP):}
The recent trend of LLP is to train a DNN using a proportion loss, originally provided by \cite{ardehaly2017co}.
The proportion loss is a bag-level cross-entropy between the correct label proportion and the predicted proportion, which is computed by averaging the probability outputs in every bag as the proportion estimation.
Many methods extend the proportion loss by introducing regularization terms or pre-training techniques \cite{ShiY2020,ardehaly2017co,DulacArnoldG2020,tsai2020,yang2021two,liu2019learning}.
In these papers, it has been reported that the accuracy decreases as the bag sizes increase.

\noindent{\bf pseudo-labeling:}
pseudo-labeling has often been used for semi-supervised learning \cite{berthelot2019mixmatch,sohn2020fixmatch}, in which a pre-trained model is first trained using few labeled data.
pseudo-labeling \cite{lee2013pseudo} assigns pseudo labels to confident unlabeled data when the maximum prediction probability estimated by a pre-trained model exceeds a threshold and re-trains the model using pseudo labels.

This pseudo-labeling is also used for several LLP methods \cite{yu2013proptosvm,DulacArnoldG2020, tokunagaECCV,ijcai2021p377}.
Yu et al. provided $\propto$-SVM, which alternately updates the pseudo labels and the SVM-based classifier. However, it can be used only for linear or kernel-based binary classification.
\cite{tokunagaECCV} tackled the LLP tasks for medical image recognition. Their proposed method generates suitable pseudo labels using several supervised instances.
\cite{DulacArnoldG2020} and \cite{ijcai2021p377} considered the hybrid method of proportion loss and pseudo-labeling. 
However, these methods degrade the performance by increasing the bag sizes.

\noindent{\bf Online decision-making for combinatorial decision space:}
Various online decision-making problems have been investigated (see, e.g., \cite{hazan2016introduction}). The task is to give a decision from the decision space sequentially with a small regret. 
Particularly, the problems for combinatorial decision space are algorithmically challenging due to the computational difficulty, and thus various problems and approaches have been proposed~\cite{rajkumar2014online,audibert2014regret,koolen2010hedging,suehiro2012online,cohen2015following}.
However, the real applications have not been studied well. 
\par
A similar study to ours is~\cite{song2020no}, where a training scheme of DNN with a noisy-labeled training set is proposed. Its approach alternately updates the decision of whether clean or noisy data and the parameters of the DNN. They utilize the online $k$-set decision framework with Follow the Perturbed Leader (FPL) algorithm~\cite{kalai2005efficient}.
However, the task is essentially different from ours, and our provided online pseudo-label decision is a more challenging problem because the decision space is a set of zero-one matrices, and thus it is difficult to utilize FPL due to the computational hardness.

\section{LLP with online pseudo-label decision}
\begin{figure}[t]
\begin{center}
   \includegraphics[width=0.7\linewidth]{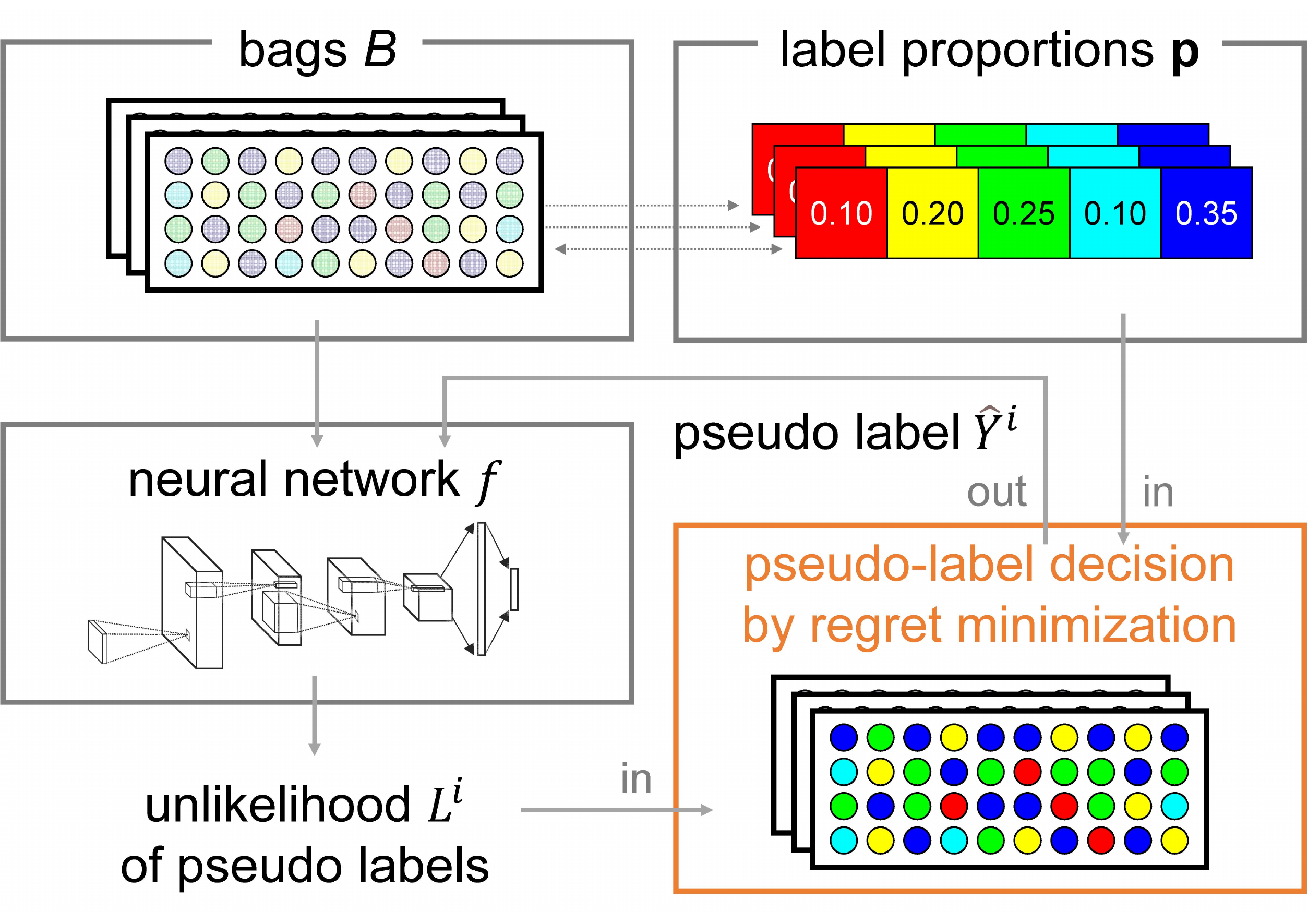}
\end{center}
\vspace{-5mm}
\caption{Overview of the proposed method, LLP with online pseudo-label decision by regret minimization.}
\label{fig:overview}
\vspace{-5mm}
\end{figure}

\vspace{-2mm}
In this section, we propose a pseudo-labeling algorithm for LLP.
The overview of the proposed method is shown in Fig.~\ref{fig:overview}.

\vspace{-1mm}
\subsection{LLP and pseudo-labeling}
In LLP, a training set contains $n$ bags, $B^1, \ldots, B^n$, and each bag $B^i$ has the set of instances, i.e., $B^i=\{x_j\}_{j=1}^{|B^i|}$.
Each $B^i$ has a label proportion $p_{c}^i = \frac{|\{j \mid j \in [|B^i|], Y^i_{c,j}=1\}|}{|B^i|}$ for any $c \in [C]$~\footnote{For a positive integer $a$, $[a]$ denotes the set $\{1,\ldots,a\}$.}, where $C$ is the number of target classes and 
$Y^i \in \{Y \mid Y \in \{0,1\}^{C\times |B^i|}, \forall j \in [|B^i|],\sum_{c=1}^{C}Y_{c,j}=1\}$
indicates \emph{unknown} labels of the instances.
The goal of the learner is to find $f$ which predicts the correct labels 
of the instances.
The problem can be considered as the optimization of not only $f$ but also the labels of instances $\hat{Y}^1, \ldots \hat{Y}^n$ according to the label proportions.
We formulate the problem of LLP as follows:
\begin{align}
\label{al:train_opt}
    \min_{\hat{Y}^1, \ldots \hat{Y}^n, f} &\sum_{i=1}^n \sum_{j =1}^{|B^i|} \ell(x^i_j, \hat{Y}^i_{:,j}, f)\\ \nonumber
    \mathrm{s. t.}~~& \forall i \in [n], \forall c \in [C],
    \frac{|\{j \mid j \in [|B^i|], \hat{Y}^i_{c,j}=1\}|}{|B^i|}=p_{c}^i,
\end{align}
where $Y_{:,j}$ denotes the $j$-th column vector of a matrix $Y$, $\ell$ is a loss function for multi-class classification.
\par
To obtain the optimal solution of the problem (\ref{alg:fpl}) is 
computationally hard. 
A straightforward way is to solve the following (i) and (ii) alternately~\cite{yu2013proptosvm};
(i) obtain $f$ for fixed pseudo labels $\hat{Y}^1, \ldots, \hat{Y}^n$,
(ii) obtain pseudo labels $\hat{Y}^1, \ldots, \hat{Y}^n$ for a fixed $f$. Then, the final $f$ and $\hat{Y}^1, \ldots, \hat{Y}^n$ are the learned model and the estimated labels, respectively.
However, when we employ a model with a high representation ability,
$f$ may overfit (possibly incorrect) initial fixed labels, and the labels are not updated.
\par
Then, we consider updating $\hat{Y}^1[t], \ldots, \hat{Y}^n[t]$ and $f[t]$ alternately at epoch $t$, where 
$\hat{Y}^i[t]$ denotes the pseudo labels of $B^i$ at epoch $t$ and $f[t]$ denotes a trained model at epoch $t$.
That is, at each epoch, we train $f[t]$ using 
pseudo labels $\hat{Y}^1[t], \ldots, \hat{Y}^n[t]$
and update the pseudo labels.
The main questions are as follows: One is how to update the pseudo labels using 
the information of label proportions and observing the behavior of $f[t]$ at each epoch.
Another is that obtaining good $\hat{Y}^1[t], \ldots, \hat{Y}^n[t]$
is computationally hard. 
While we can efficiently obtain an optimal $\hat{Y}^1[t], \ldots, \hat{Y}^n[t]$ by greedy algorithm in binary classification case (see, e.g., \cite{yu2013proptosvm}),
the optimization problem becomes a Mixed Integer Problem (MIP), which is an NP-complete problem in multi-class cases.
Therefore, pseudo-labeling for LLP is a simple but challenging approach.

\subsection{Proposed procedure}
$\mathcal{Y}^i$ denotes the decision space of $Y^i$ for any $i \in [n]$, i.e.,
$\mathcal{Y}^i = \{Y \mid Y \in \{0,1\}^{C\times |B^i|}, \forall j \in [|B^i|],\sum_{c=1}^{C}Y_{c,j}=1,~\mathrm{and}~ \forall c \in [C], \sum_{j=1}^{|B^i|}Y_{c,j} = k_{c}^i\}$, where $k_{c}^i=|B^i|p^i_c$ (i.e., the number of instances belonging to class $c$ in a bag $B^i$).
For any $B^i$, we define an ``unlikelihood'' of pseudo-labeling
to the instances as $L^i \in [0,1]^{C\times |B^i|}$. 
For example, if $c$ is not likely as a pseudo label of $x^i_j$, $L^i_{c,j}$ takes a higher value (detailed later).\par
We provide a 3-step procedure for LLP with pseudo-labeling as follows:
Let $\hat{Y}^i[1] \in \mathcal{Y}^i$ $(i \in [n])$ be initial pseudo labels and $f[0]$ be an initial DNN. At each epoch $t = 1, \ldots, T$, for any $i\in [n]$,
\begin{enumerate}
    \item Obtain $f[t]$ by training $f[t-1]$ using $\ell$ and the pseudo-labeled instances $((x^1_1, \hat{Y}^1_{:,1}[t]), \ldots, (x^n_{|B^n|},\hat{Y}^n_{:,|B^n|}[t]))$.
    \item Obtain unlikelihood $L^i[t]$. 
    \item Decide a next $\hat{Y}^i[t]$ by observing $L^i[1], \ldots, L^i[t]$.
\end{enumerate}
\par
In this paper, we compute the unlikelihood at each epoch $t$ as
\begin{align}
\label{align:unlikelihood}
L^i_{c,j}[t] =
  \begin{cases}
     1 - \mathrm{conf}(x^i_j, c, f[t])  &\hat{Y}_{c,j}^i[t]=1 \\
    {\displaystyle \max_{c \in [C]}}~\mathrm{conf}(x^i_j, c, f[t]) - \mathrm{conf}(x^i_j, c, f[t])   &\text{otherwise}, 
  \end{cases}
\end{align}
where $\mathrm{conf}$ returns the confidence of $f$ when $x^i_j$ is assigned to class $c$ (i.e., posterior probability).
The motivation of this unlikelihood is rather simple.
In the first case, if $f[t]$ learned $x_j^i$ with the pseudo label $c$, and $x_j^i$ is assigned to $c$ by $f[t]$ with high confidence, $c$ is likely to be a correct label of $x_j^i$. On the other hand, if $x_j^i$ is assigned to $c$ by $f[t]$ with low confidence, $c$ is not likely to be a correct label.
In the second case, if $f[t]$ learned $x_j^i$ with some pseudo label other than $c$, and $x_j^i$ is assigned to $c$ by $f[t]$ with high confidence relative to the maximum confidence, $c$ is likely to be a correct label of $x_j^i$.
More specifically, if it is difficult to learn $x_j^i$ with the pseudo label $c$ (i.e., $\max_{c \in [C]}~\mathrm{conf}(x^i_j, c, f[t])$ is small), 
and $x_j^i$ is assigned to $c$ by $f[t]$ with similar confidence to the maximum confidence, $L^i_{c,j}[t]$ becomes a low value.
\par

\subsection{pseudo-labeling with regret minimization approach}
We consider deciding the pseudo labels for each bag $B^i$ individually.
Since the unlikelihood of the decided pseudo label for $x^i_j$ at epoch $t$ can be formulated as  $\hat{Y}_{:,j}^{i}[t]^\top L_{:,j}^i[t]$,
we can evaluate the performance of the decided pseudo labels $\hat{Y}^{i}[t]$ by the total unlikelihood over $B^i$ : $\sum_{j=1}^{|B^i|}\hat{Y}_{:,j}^{i}[t]^\top L_{:,j}^i[t]$. 
Therefore, a straightforward goal is to predict $\hat{Y}^i[t]$ which minimizes $\sum_{j=1}^{|B^i|}\hat{Y}_{:,j}^{i}[t]^\top L_{:,j}^i[t]$ at each epoch.
However, $L^i[t]$ is revealed after the training (step 2) and it is difficult to give such $\hat{Y}^i[t]$ in step 1. 
Moreover, due to the instability of DNN training (especially in early epochs), 
$L^i[t]$ may fluctuate and thus to predict $\hat{Y}^i[t]$ which minimizes $\sum_{j=1}^{|B^i|}\hat{Y}_{:,j}^{i}[t]^\top L_{:,j}^i[t]$ is not a reasonable goal.
Then, we aim to give suitable pseudo labels averagely over epochs:
\begin{align}
    (\hat{Y}^i)^* = \argmin_{\hat{Y}^i \in \mathcal{Y}^i} \sum_{t=1}^T \sum_{j=1}^{|B^i|}\hat{Y}_{:,j}^{i\top} L_{:,j}^i[t].
\end{align}
\vspace{2mm}
The optimization problem is still difficult because we need to decide $(\hat{Y}^i)^*$ online, and the best solution $(\hat{Y}^i)^*$ can be revealed after $T$ epochs.
Therefore, we consider deciding $\hat{Y}^{i}[t]$ online to minimize the \emph{regret} for each bag $B^i$, which is defined as:
\begin{align}
    \label{align:regret}
    R_{T}^i = 
    \sum_{t=1}^T \sum_{j=1}^{|B^i|}\hat{Y}_{:,j}^{i}[t]^\top L_{:,j}^i[t]
    - \sum_{t=1}^T \sum_{j=1}^{|B^i|}(\hat{Y}_{:,j}^i)^{*\top} L_{:,j}^i[t].
\end{align}
\vspace{2mm}
The regret measure is used in online decision-making, which evaluates the difference in the relative performance between the actual decisions and the best decision in hindsight.
That is, to achieve small regret indicates that the performance of the actual decisions is competitive to the best decision.
\par

The most significant advantage of our online pseudo-labeling decision is that
we can have theoretical support on the regret under any tough situation. As aforementioned, $L^i[t]$ may fluctuate during DNN training. However, as detailed later, by utilizing a regret-bounded scheme, we can guarantee the performance of the pseudo labels for any sequences of $L^i[1], \allowbreak \ldots, \allowbreak L^i[T]$, i.e., we do not need to care about the fluctuation of the DNN.
Thus, we can decide on likely pseudo labels online by the regret minimization approach.
\par
\subsection{pseudo-label decision using Follow the Perturbed Leader (FPL)}
To minimize Eq.~(\ref{align:regret}) online, we employ FPL~\cite{kalai2005efficient}, a popular regret minimization algorithm.
The details of our algorithm using FPL are shown in Algorithm~\ref{alg:fpl}.
The remarkable feature of FPL is to add the perturbation $Z^i \in \mathbb{R}^{C \times|B^i|}$ with the rate $\eta$ to the original $L^i[t]$ as shown in line 8 and 9, where $\eta$ is the hyperparameter which controls the effect of the perturbation. If we naively use the optimal decision without perturbation, the decision is the optimal pseudo labels only at epoch $t$, and thus it may overfit to the fluctuated $L^i[t]$.
\par
Theoretically, the perturbation allows us to avoid such overfitting. Using the analysis of FPL~\cite{cohen2015following}, for any sequences $L^i[1], \ldots,\allowbreak L^i[T]$, we can guarantee the upper bound of the regret as $\mathbb{E}[R^i_{T}] = O(|B^i|\sqrt{T \ln |\mathcal{Y}^i|})$, where the expectation is derived from the randomness of FPL. This bound indicates that we can logarithmically suppress the complexity of the combinatorially large decision space $|\mathcal{Y}_i|$, and the regret converges with increasing the epochs.
\begin{algorithm}[t]
\caption{pseudo-label decision by regret minimization.}
\label{alg:fpl}
\begin{algorithmic}[1]
\Inputs{Training bags $(B^1, \bp^1), \ldots, (B^n, \bp^n)$, total epochs $T$, initial DNN $f[0]$, loss $\ell$, $\eta > 0$}
\Outputs{$f[T]$: trained DNN}
\Initialize{$\forall i \in [n]$, $\hat{Y}^i[1] \in \mathcal{Y}^i$ and $Z^i \in \mathbb{R}^{C \times |B^i|}$}
\For {epochs $t=1, \ldots, T$}
  \State \multiline{Obtain $f[t]$ by training $f[t-1]$ 
   using $\ell$ and the pseudo-labeled instances $((x^1_1, \hat{Y}^1_{:,1}[t]), \ldots, (x^n_{|B^n|},\hat{Y}^n_{:,|B^n|}[t])) $.}
  \For {$i=1, \ldots, n$}
  \State{Obtain $L^i[t]$ by Eq.(\ref{align:unlikelihood}).}
      \State \multiline{Sample the perturbation ${Z}_{c,j}^i \sim \mathcal{N}(0,1)$ for any $c\in [C]$ and $j \in [|B^i|]$.}
    \State \multiline{Decision pseudo labels by 
    \vspace{-0.1cm}
    \begin{align}
     \label{align:offline_opt}
         \hat{Y}^i[t+1] = \arg\min_{\hat{Y}^i \in \mathcal{Y}^i}
         \left(\sum_{\tau=1}^t \sum_{j=1}^{|B^i|}\hat{Y}_{:,j}^{i\top} (L_{:,j}^i[t] + \eta Z_{:,j}^i) \right)
     \end{align}
    }
             \vspace{-0.7cm}
  \EndFor
\EndFor
\end{algorithmic}
\end{algorithm}

\par
The remaining issue is how to obtain the solution of 
Eq.~(\ref{align:offline_opt}), which is explicitly formulated as follows.
\begin{align}
\label{align:mip}
\min_{\hat{Y}^i \in \{0,1\}^{C\times |B^i|}}
        &
        \sum_{\tau=1}^t \sum_{j=1}^{|B^i|} \left(\hat{Y}_{:,j}^{i}[t]^\top L_{:,j}^i[t] + \eta \hat{Y}_{:,j}^{i\top} Z_{:,j}^i\right) 
        \\ \nonumber
\mathrm{s. t.} \quad \quad &\forall j \in [|B^i|],\sum_{c=1}^{C}\hat{Y}_{c,j}^i=1,
 ~~\forall c \in [C], \sum_{j=1}^{|B^i|}\hat{Y}_{c,j}^i = k_{c}^i.
\end{align}
The optimization problem is MIP, and it is NP-complete~\cite{papadimitriou1998combinatorial} in general.
However, the constraint matrix is totally unimodular; thus   
, we can obtain the optimal solution in polynomial time by relaxing to the linear programming problem.


\section{Experiments}

As we introduced, our focus is LLP with large bag sizes.
Following previous LLP research ~\cite{ShiY2020,DulacArnoldG2020,tsai2020,yang2021two,liu2019learning}, we consider virtual LLP using SVHN and CIFAR-10 datasets.
First, we show the results on LLP with large bag sizes  \cite{krizhevsky2009learning} compared with the state-of-the-art methods that use the proportion loss. Second, we show the ablation study of our online pseudo-labeling approach.

\subsection{Comparative methods}
\paragraph{Methods using proportion loss:}
As a standard baseline, we consider a DNN trained with the proportion loss (we call the method \textbf{PL} for short).
A standard proportion loss is formulated as below:
\begin{align}
    \ell_{\mathrm{prop}}(B^i, \bp^i, f) = - \sum_{c=1}^{C} p_c^i \log \frac{1}{|B^i|} \sum_{j=1}^{|B^i|} \mathrm{conf}(x_j^i, c, f).
\end{align}
We also compare with \textbf{$\Pi$-model}~\cite{laine2017} and \textbf{LLP-VAT}~\cite{tsai2020}, which are the state-of-the-art LLP methods using the proportion loss, the implementations of which are publicly available.
\par
\paragraph{Methods for an ablation study:}
First, to show the proposed unlikelihood (see Eq.~(\ref{align:unlikelihood})) is effective, we compare it with the simpler likelihood as below:
\begin{align}
    \label{align:unlikelihood2}
    L^i_{c,j}[t] = 1 - \mathrm{conf}(x^i_j, c, f[t]).
\end{align}

Second, to evaluate the effectiveness of our proposed pseudo-labeling approach based on the regret minimization, we also compare it with the following two methods.
One is ``Greedy,'' which does not use the perturbation term to decide the pseudo labels in Eq.~(\ref{align:offline_opt}).
Another is ``Naive'' which naively update the pseudo labels only using the latest $L^i[t]$, i.e., $\hat{Y}^i[t+1] = \arg\min_{\hat{Y}^i \in \mathcal{Y}^i} \sum_{j=1}^{|B^i|}\hat{Y}_{:,j}^{i\top}L_{:,j}^i[t]$.
We used Eq.~(\ref{align:unlikelihood}) as the unlikelihood for Greedy and Naive.

\subsection{Implementation details}
For all methods, we used ResNet18. 
The learning rate was set to $0.0003$, and the model was optimized by Adam~\cite{kingma2014adam}.
The number of training epochs was fixed at $400$.
The mini-batch size (number of bags) was fixed to 4. 
The hyperparameter $\eta$ of the proposed method is set to $5$.
The number of original training instances was fixed to $102400$.
Training instances were separated at $7:3$ for training and validation. We randomly separated the original training instances into bags. The bag sizes (i.e., the numbers of instances in a bag) were $64, 128, 256, 512, 1024, 2048, 4096$.
For example, if the bag size is $1024$, the number of bags ($n$) is $100$. We randomly created proportions $n$ times, and then the instances of the bags were chosen based on each proportion. 
Including the proposed method, the best model for evaluation was chosen based on the mean absolute label-proportion error of the validation set.

\subsection{Results of comparative experiments}

\begin{table}[t]
\centering
\caption{Accuracy (\%) on CIFAR-10.}
\label{tab:quantitative_result_cifar10}
\scriptsize
\begin{tabular}{lccccccc}
\hline
             & \multicolumn{7}{c}{bag size (number of bags)} \\ \cline{2-8}
      & 64 & 128 & 256 & 512 & 1024 & 2048 & 4096 \\ 
method      & (1600) & (800) & (400) & (200) & (100) & (50) & (25) \\ \hline
PL          & \textbf{61.24}  &  57.12 & 55.54  & 55.07 & 51.09 & 50.32  & 42.60  \\
$\Pi$-model &  60.68  &  55.97   & 52.10    &  51.56   &   50.03   &  47.96    &  47.15    \\
LLP-VAT & 59.23   & 53.64    & 52.88    &  50.52   & 50.25     & 45.53     & 44.81     \\
\hline
ours   & 58.59 &  \textbf{59.34}  & \textbf{60.76}    & \textbf{61.13}    & \textbf{61.24}    & \textbf{59.83}     & \textbf{59.86}      \\ 
\hline
\end{tabular}

\end{table}
\begin{table}[t]
\centering
\caption{Accuracy (\%) on SVHN.}
\label{tab:quantitative_result_svhn}
\scriptsize
\begin{tabular}{lccccccc}
\hline
             & \multicolumn{7}{c}{bag size (number of bags)} \\ \cline{2-8}
              & 64 & 128 & 256 & 512 & 1024 & 2048 & 4096 \\ 
method      & (1600) & (800) & (400) & (200) & (100) & (50) & (25) \\ \hline
PL          & {90.19}  &  \textbf{87.99} & \textbf{87.19}  & \textbf{87.35} &  84.76 &  81.65 & 78.85  \\
$\Pi$-model &  \textbf{90.94}  & 87.08    &  82.97   &  81.87   & 77.27     & 79.05     &  77.58   \\
LLP-VAT & 88.02   & 84.97    & 83.04    &  81.96   &  80.09    &  80.17    & 78.58    \\
\hline
ours & 87.36 & 85.42   & 84.79  &  85.87   & \textbf{85.99}    & \textbf{86.08}     & \textbf{86.37}      \\  
\hline
\end{tabular}
\end{table}

\begin{figure}[t]
\vspace{3mm}
\begin{center}
\subfigure[CIFAR-10]{
\includegraphics[scale=0.3]{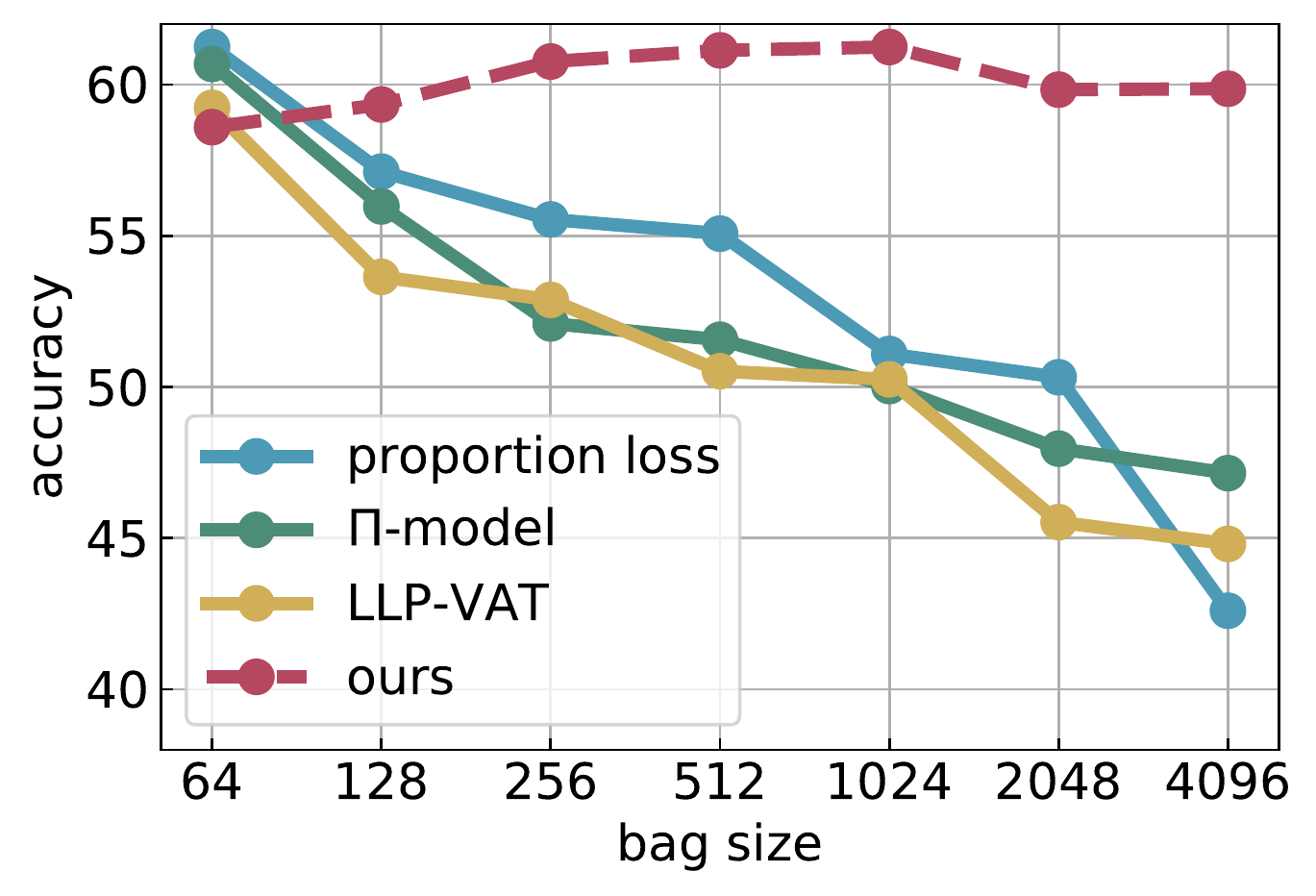}}
\subfigure[SVHN]{
\includegraphics[scale=0.3]{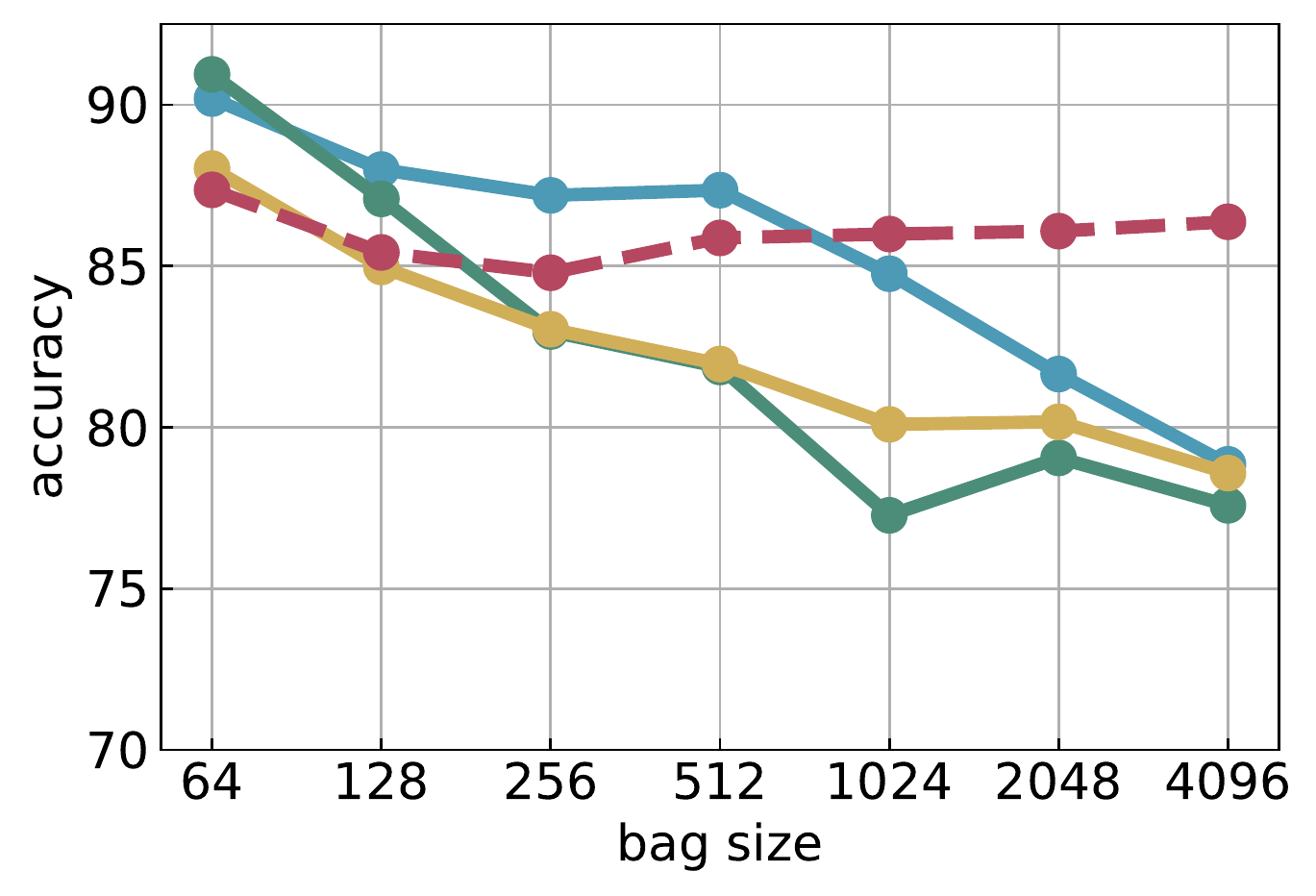}}
\end{center}
\vspace{-6mm}
\caption{The accuracy at different bag sizes.}
\label{fig:acc}
\end{figure}

Tables~\ref{tab:quantitative_result_cifar10} and \ref{tab:quantitative_result_svhn} show the results on CIFAR-10 and SVHN, respectively. 
We can see that our method achieves the best accuracy when the bag size is large on both datasets.
Fig.~\ref{fig:acc} plots the accuracies at different bag sizes. The accuracy of the methods using proportion loss was degraded by larger bag sizes. On the other hand, our method achieved high accuracy stably even when the bag size was large.
We can say that the proposed method is robust to increasing the bag sizes.
\par


\subsection{Ablation study}

\begin{table}[t]
\centering
\caption{Accuracy (\%) on CIFAR-10 (ablation study).}
\label{tab:ablation}
\scriptsize
\begin{tabular}{lccccccc}
\hline
             & \multicolumn{7}{c}{bag size (number of bags)} \\ \cline{2-8}
      & 64 & 128 & 256 & 512 & 1024 & 2048 & 4096 \\ 
method     & (1600) & (800) & (400) & (200) & (100) & (50) & (25) \\ \hline
ours w/ Eq.~(\ref{align:unlikelihood})  & \textbf{58.59} &  \textbf{59.34}  & \textbf{60.76}    & \textbf{61.13}    & \textbf{61.24}    & \textbf{59.83}     & \textbf{59.86}      \\ 
ours w/ Eq.~(\ref{align:unlikelihood2})  & 55.01 & 53.18   &  54.11   & 52.80    &  53.05   & 52.48     & 50.75 \\
Greedy       & 25.77   &  22.51   &  22.64   & 25.45    & 23.82     & 22.81     & 21.35     \\
Naive      &  39.05  & 35.47    &  36.22   & 36.44    & 35.33     &   32.86   &  32.76    \\   
\hline
\end{tabular}
\end{table}

\begin{figure}[t]
\vspace{3mm}
\begin{center}
\includegraphics[scale=0.3]{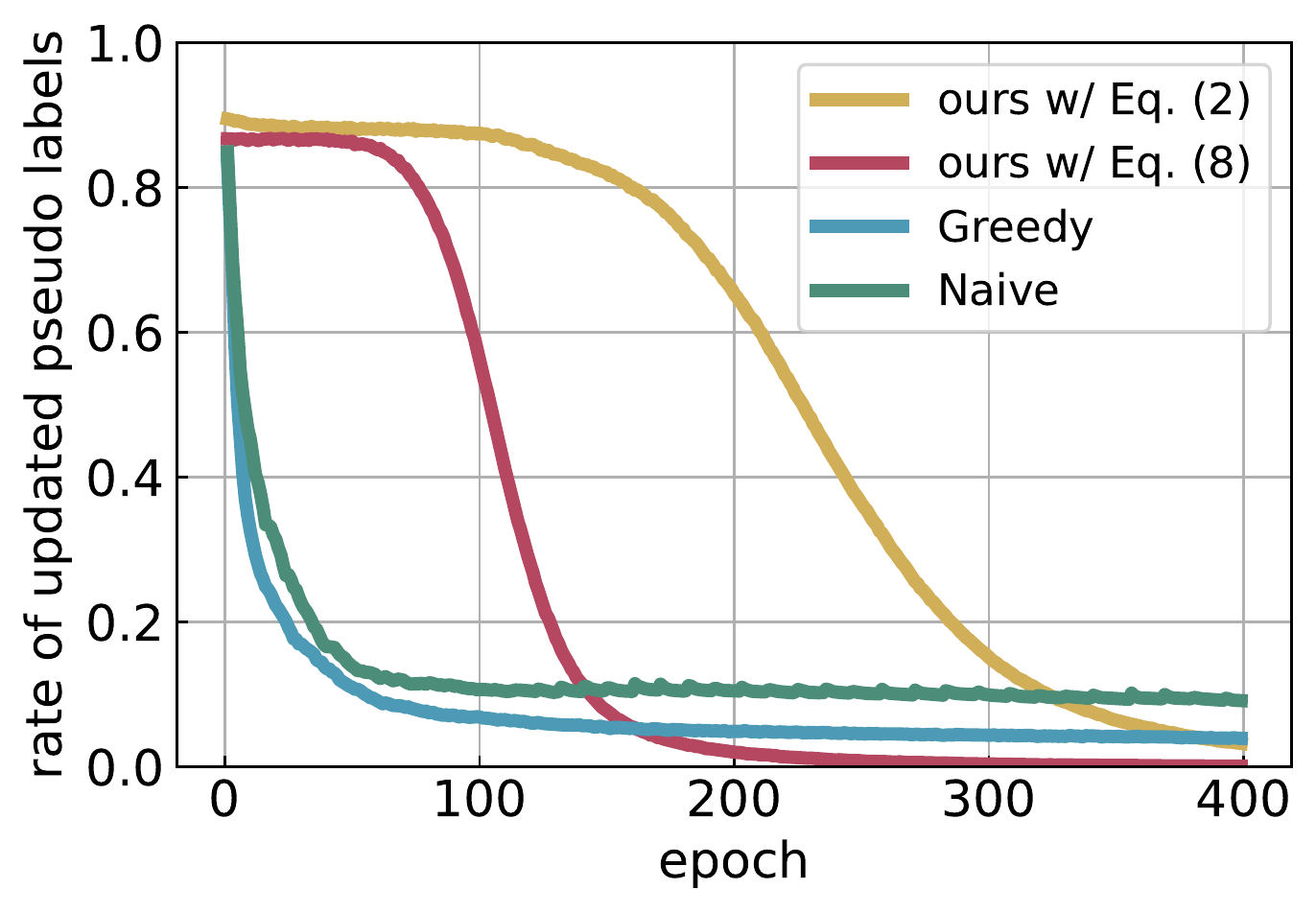}
\includegraphics[scale=0.3]{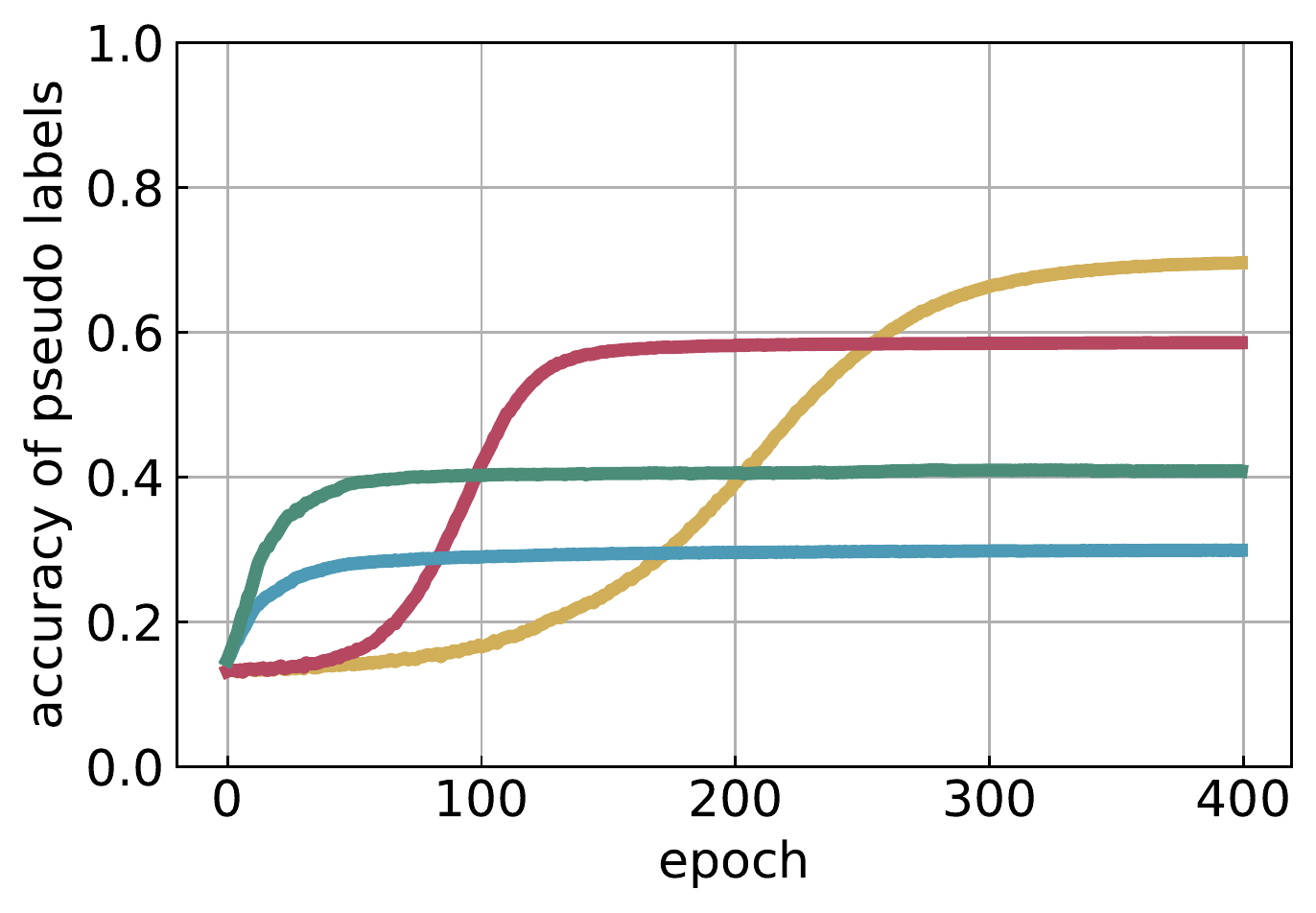}
\end{center}
\vspace{-6mm}
\caption{(Left) The rate of the updated pseudo labels compared to the previous epoch on CIFAR-10 with bag size 4096. (Right) The accuracy of the pseudo labels.}
\label{fig:acc_p_label}
\end{figure}

As shown in Table~\ref{tab:ablation}, the performances of Greedy and Naive approaches were significantly worse than our proposed method. Moreover, we can see that the unlikelihood Eq.~(\ref{align:unlikelihood}) performed better than Eq.~(\ref{align:unlikelihood2}). The results indicate that the regret minimization approach with the unlikelihood evaluation by Eq.~(\ref{align:unlikelihood}) effectively works for the pseudo-labeling.
\par
In Fig.~\ref{fig:acc_p_label}, we can observe the difference in the pseudo-labeling results between ours and others. 
The left side of Fig.~\ref{fig:acc_p_label} shows how much the pseudo labels have been updated by epochs on CIFAR-10. Naive and no perturbation approaches fixed most of the pseudo labels at the initial 5 epochs, and the accuracies were not improved much at the later epochs as shown in the right side of Fig.~\ref{fig:acc_p_label}.
On the other hand, the proposed method updated more pseudo labels than Greedy and Naive. This is because the effect of perturbation is larger than the effect of the original unlikelihood $\sum_{t} \hat{Y}^{i\top}L^i[t]$ at early epochs. That is, the proposed method can explore various pseudo labels and achieve better performance.

\section{Conclusion}
\label{sec:conc}
In this paper, we propose a novel LLP method based on pseudo-labeling with regret minimization, which is robust to increasing the bag sizes compared to the previous LLP methods.
The proposed method is that, by assigning the pseudo labels to the instances over the bags, we can make full use of instances to train a model even if the number of bags is small.
We demonstrated the effectiveness of the proposed method through comparative and ablation studies.
\newpage

\bibliographystyle{IEEEbib}
\bibliography{main}

\clearpage
\begin{center}
    \textbf{Appendix}
\end{center}

\begin{table}[th]
\centering
\caption{Accuracy (\%) on SpokenArabicDigits.}
\vspace{2mm}
\label{tab:acc}
\begin{tabular}{lccccc}
\hline
             & \multicolumn{5}{c}{bag size } \\ \cline{2-6}
method      & 64 & 128 & 256 & 512 & 1024  \\ \hline
PL          & 98.26& 98.32& 97.39& 96.44& 96.42  \\
$\Pi$-model &  98.51& 98.39& 97.71& 97.05& 97.01     \\
LLP-VAT & \textbf{98.64}& 98.50& 97.73& 97.57& 96.42        \\
\hline
ours   & 98.51& \textbf{98.62}& \textbf{98.74}& \textbf{98.23}& \textbf{98.36}      \\   
\hline
\end{tabular}
\end{table}
\begin{figure}[th]
\vspace{3mm}
\begin{center}
\includegraphics[scale=0.4]{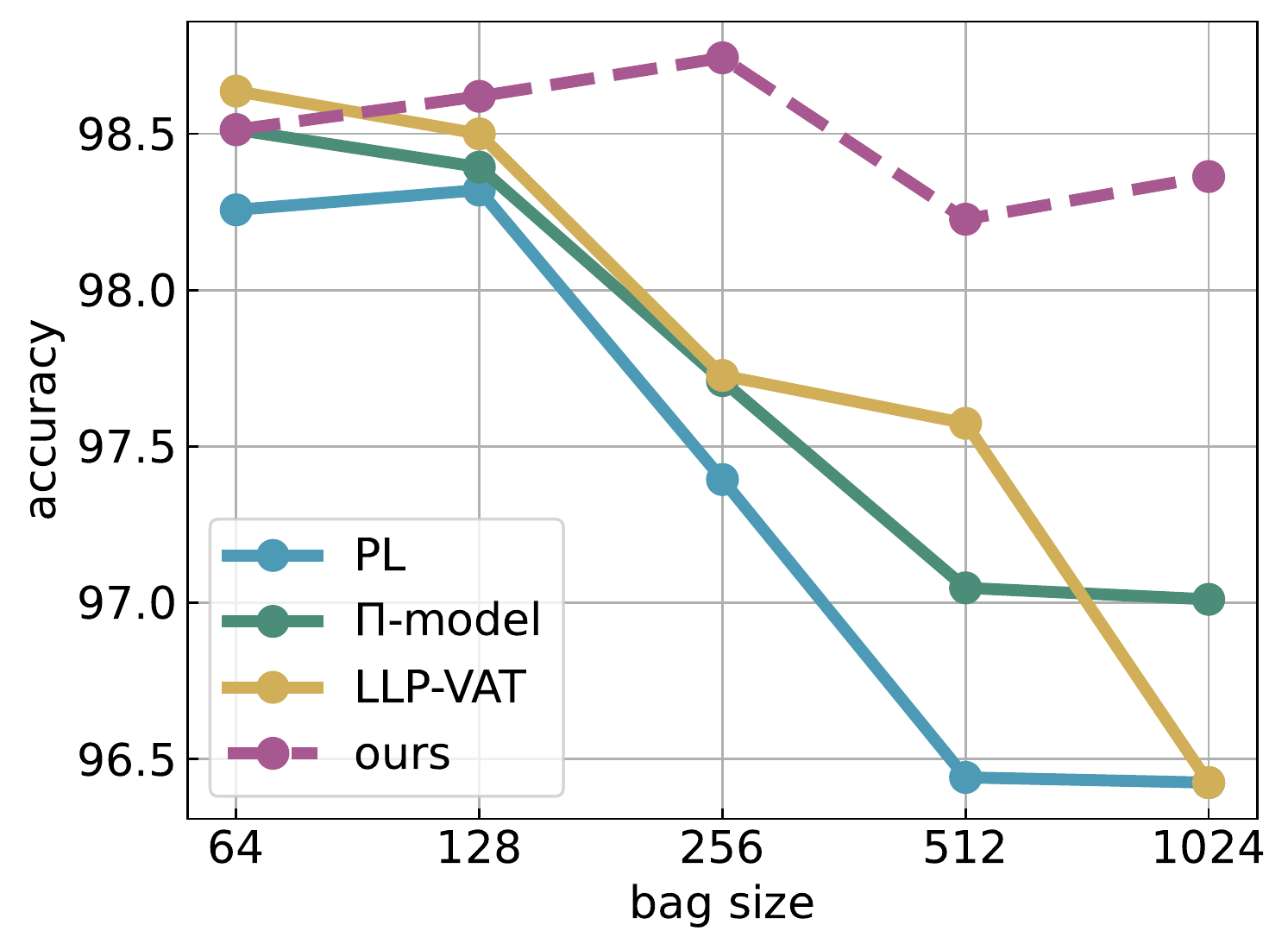}
\end{center}
\vspace{-8mm}
\caption{The accuracy at different bag sizes.}\vspace{-3mm}
\label{fig:acc}
\vspace{8mm}
\end{figure}

As an application example for signal processing, we conducted an additional experiment on a {\em speech signal dataset}. We used SpokenArabicDigits from the University of California Riverside(UCR) Archive\footnote{{\tt https://www.cs.ucr.edu/\textasciitilde eamonn/time\_series\_data\_2018/}.}.
Following the experiment setup in our main paper,
we made the virtual LLP datasets using each speech data as an instance. The bag sizes (i.e., the numbers of instances in a bag) were $64, 128, 256, 512, 1024$. The technical difference from the experiments in the main paper is only to use 1D convolution layers instead of 2D.
Other implementation details are the same as the other experiments.

The results are shown in Table~\ref{tab:acc} and Fig.~\ref{fig:acc}.
Similar to the CIFAR10 and SVHN results, the accuracy of the comparative methods with the standard proportion loss in UCR was degraded with increasing the bag sizes. On the other hand, our method achieved high accuracy at arbitrary bag sizes. From the results, we can say that our method is also applicable to LLP tasks with signal data.
Of course, the theoretical supports (upper bound of the regret, efficiency of the algorithm) are still valid not only images but also signals, and any other data. 

\end{document}